# Towards String-to-Tree Neural Machine Translation

**Roee Aharoni & Yoav Goldberg**
Computer Science Department
Bar-Ilan University
Ramat-Gan, Israel
{roee.aharoni,yoav.goldberg}@gmail.com

## Abstract

We present a simple method to incorporate syntactic information about the target language in a neural machine translation system by translating into linearized, lexicalized constituency trees. Experiments on the WMT16 German-English news translation task shown improved BLEU scores when compared to a syntax-agnostic NMT baseline trained on the same dataset. An analysis of the translations from the syntax-aware system shows that it performs more reordering during translation in comparison to the baseline. A small-scale human evaluation also showed an advantage to the syntax-aware system.

## 1 Introduction and Model

Neural Machine Translation (NMT) (Kalchbrenner and Blunsom, 2013; Sutskever et al., 2014; Bahdanau et al., 2014) has recently became the state-of-the-art approach to machine translation (Bojar et al., 2016), while being much simpler than the previously dominant phrase-based statistical machine translation (SMT) approaches (Koehn, 2010). NMT models usually do not make explicit use of syntactic information about the languages at hand. However, a large body of work was dedicated to syntax-based SMT (Williams et al., 2016). One prominent approach to syntax-based SMT is string-to-tree (S2T) translation (Yamada and Knight, 2001, 2002), in which a source-language string is translated into a target-language tree. S2T approaches to SMT help to ensure the resulting translations have valid syntactic structure, while also mediating flexible reordering between the source and target languages. The main formalism driving current S2T SMT systems is GHKM rules (Galley et al., 2004, 2006), which are synchronous transduction grammar (STSG) fragments, extracted from word-aligned sentence pairs with syntactic trees on one side. The GHKM translation rules allow flexible reordering on all levels of the parse-tree.

**We suggest** that NMT can also benefit from the incorporation of syntactic knowledge, and propose a simple method of performing string-to-tree neural machine translation. Our method is inspired by recent works in syntactic parsing, which model trees as sequences (Vinyals et al., 2015; Choe and Charniak, 2016). Namely, we translate a source sentence into a linearized, lexicalized constituency tree, as demonstrated in Figure 2. Figure 1 shows a translation from our neural S2T model compared to one from a vanilla NMT model for the same source sentence, as well as the attention-induced word alignments of the two models.

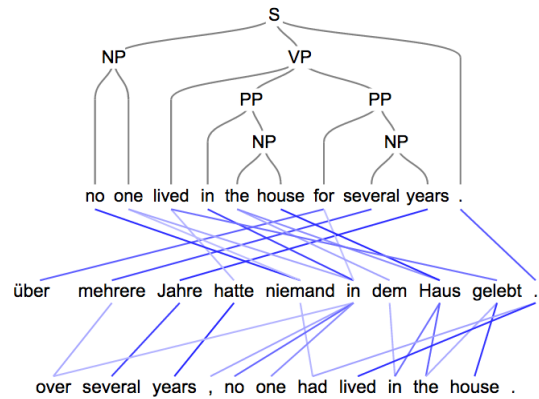

Figure 1: Top - a lexicalized tree translation predicted by the bpe2tree model. Bottom - a translation for the same sentence from the bpe2bpe model. The blue lines are drawn according to the attention weights predicted by each model.

Note that the linearized trees we predict are different in their structure from those in Vinyals et al. (2015) as instead of having part of speech tags as terminals, they contain the words of the translated sentence. We intentionally omit the POS information as including it would result in significantly

> **Jane hatte eine Katze .** → ($_{ROOT}$ ($_S$ ($_{NP}$ **Jane** )$_{NP}$ ($_{VP}$ **had** ($_{NP}$ **a cat** )$_{NP}$ )$_{VP}$ **.** )$_S$ )$_{ROOT}$ longer sequences. The S2T model is trained on parallel corpora in which the target sentences are automatically parsed. Since this modeling keeps the form of a sequence-to-sequence learning task, we can employ the conventional attention-based sequence to sequence paradigm (Bahdanau et al., 2014) as-is, while enriching the output with syntactic information.

Figure 2: An example of a translation from a string to a linearized, lexicalized constituency tree.

**Related Work** Some recent works did propose to incorporate syntactic or other linguistic knowledge into NMT systems, although mainly on the source side: Eriguchi et al. (2016a,b) replace the encoder in an attention-based model with a Tree-LSTM (Tai et al., 2015) over a constituency parse tree; Bastings et al. (2017) encoded sentences using graph-convolutional networks over dependency trees; Sennrich and Haddow (2016) proposed a factored NMT approach, where each source word embedding is concatenated to embeddings of linguistic features of the word; Luong et al. (2015) incorporated syntactic knowledge via multi-task sequence to sequence learning: their system included a single encoder with multiple decoders, one of which attempts to predict the parse-tree of the source sentence; Stahlberg et al. (2016) proposed a hybrid approach in which translations are scored by combining scores from an NMT system with scores from a Hiero (Chiang, 2005, 2007) system. Shi et al. (2016) explored the syntactic knowledge encoded by an NMT encoder, showing the encoded vector can be used to predict syntactic information like constituency trees, voice and tense with high accuracy.

In parallel and highly related to our work, Eriguchi et al. (2017) proposed to model the target syntax in NMT in the form of dependency trees by using an RNNG-based decoder (Dyer et al., 2016), while Nadejde et al. (2017) incorporated target syntax by predicting CCG tags serialized into the target translation. Our work differs from those by modeling syntax using constituency trees, as was previously common in the "traditional" syntax-based machine translation literature.

## 2 Experiments & Results

**Experimental Setup** We first experiment in a resource-rich setting by using the German-English portion of the WMT16 news translation task (Bojar et al., 2016), with 4.5 million sentence pairs. We then experiment in a low-resource scenario using the German, Russian and Czech to English training data from the News Commentary v8 corpus, following Eriguchi et al. (2017). In all cases we parse the English sentences into constituency trees using the BLLIP parser (Charniak and Johnson, 2005).[1] To enable an open vocabulary translation we used sub-word units obtained via BPE (Sennrich et al., 2016b) on both source and target.[2]

In each experiment we train two models. A **baseline** model (bpe2bpe), trained to translate from the source language sentences to English sentences without any syntactic annotation, and a **string-to-linearized-tree** model (bpe2tree), trained to translate into English linearized constituency trees as shown in Figure 2. Words are segmented into sub-word units using the BPE model we learn on the raw parallel data. We use the NEMATUS (Sennrich et al., 2017)[3] implementation of an attention-based NMT model.[4] We trained the models until there was no improvement on the development set in 10 consecutive checkpoints. Note that the only difference between the baseline and the bpe2tree model is the syntactic information, as they have a nearly-identical amount of model parameters (the only additional parameters to the syntax-aware system are the embeddings for the brackets of the trees).

For all models we report results of the best performing single model on the dev-set (newstest2013+newstest2014 in the resource rich setting, newstest2015 in the rest, as measured by BLEU) when translating newstest2015 and newstest2016, similarly to Sennrich et al. (2016a); Eriguchi et al. (2017). To evaluate the string-to-tree translations we derive the surface form by removing the symbols that stand for non-terminals in the tree, followed by merging the sub-words. We also report the results of an ensemble of the last 5 checkpoints saved during each model training. We compute BLEU scores using the `mteval-v13a.pl` script from the Moses toolkit (Koehn et al., 2007).

---

[1] https://github.com/BLLIP/bllip-parser
[2] https://github.com/rsennrich/subword-nmt
[3] https://github.com/rsennrich/nematus
[4] Further technical details of the setup and training are available in the supplementary material.

| system | newstest2015 | newstest2016 |
|---|---|---|
| bpe2bpe | 27.33 | 31.19 |
| bpe2tree | 27.36 | 32.13 |
| bpe2bpe ens. | 28.62 | 32.38 |
| bpe2tree ens. | 28.7 | 33.24 |

Table 1: BLEU results for the WMT16 experiment

**Results** As shown in Table 1, for the resource-rich setting, the single models (bpe2bpe, bpe2tree) perform similarly in terms of BLEU on newstest2015. On newstest2016 we witness an advantage to the bpe2tree model. A similar trend is found when evaluating the model ensembles: while they improve results for both models, we again see an advantage to the bpe2tree model on newstest2016. Table 2 shows the results in the low-resource setting, where the bpe2tree model is consistently better than the bpe2bpe baseline. We find this interesting as the syntax-aware system performs a much harder task (predicting trees on top of the translations, thus handling much longer output sequences) while having a nearly-identical amount of model parameters. In order to better understand where or how the syntactic information improves translation quality, we perform a closer analysis of the WMT16 experiment.

## 3 Analysis

**The Resulting Trees** Our model produced valid trees for 5970 out of 6003 sentences in the development set. While we did not perform an in-depth error-analysis, the trees seem to follow the syntax of English, and most choices seem reasonable.

**Quantifying Reordering** English and German differ in word order, requiring a significant amount of reordering to generate a fluent translation. A major benefit of S2T models in SMT is facilitating reordering. Does this also hold for our neural S2T model? We compare the amount of reordering in the bpe2bpe and bpe2tree models using a distortion score based on the alignments derived from the attention weights of the corresponding systems. We first convert the attention weights to hard alignments by taking for each target word the source word with highest attention weight. For an $n$-word target sentence $t$ and source sentence $s$ let $a(i)$ be the position of the source word aligned to the target word in position

| | system | newstest2015 | newstest2016 |
|---|---|---|---|
| DE-EN | bpe2bpe | 13.81 | 14.16 |
| | bpe2tree | 14.55 | 16.13 |
| | bpe2bpe ens. | 14.42 | 15.07 |
| | bpe2tree ens. | 15.69 | 17.21 |
| RU-EN | bpe2bpe | 12.58 | 11.37 |
| | bpe2tree | 12.92 | 11.94 |
| | bpe2bpe ens. | 13.36 | 11.91 |
| | bpe2tree ens. | 13.66 | 12.89 |
| CS-EN | bpe2bpe | 10.85 | 11.23 |
| | bpe2tree | 11.54 | 11.65 |
| | bpe2bpe ens. | 11.46 | 11.77 |
| | bpe2tree ens. | 12.43 | 12.68 |

Table 2: BLEU results for the low-resource experiments (News Commentary v8)

$i$. We define:

$$d(s,t) = \frac{1}{n}\sum_{i=2}^{n} |a(i) - a(i-1)|$$

For example, for the translations in Figure 1, the above score for the bpe2tree model is 2.73, while the score for the bpe2bpe model is 1.27 as the bpe2tree model did more reordering. Note that for the bpe2tree model we compute the score only on tokens which correspond to terminals (words or sub-words) in the tree. We compute this score for each source-target pair on newstest2015 for each model. Figure 3 shows a histogram of the binned score counts. The bpe2tree model has more translations with distortion scores in bins 1-onward and significantly less translations in the least-reordering bin (0) when compared to the bpe2bpe model, indicating that the syntactic information encouraged the model to perform more reordering.[5] Figure 4 tracks the distortion scores throughout the learning process, plotting the average dev-set scores for the model checkpoints saved every 30k updates. Interestingly, both models obey to the following trend: open with a relatively high distortion score, followed by a steep decrease, and from there ascend gradually. The bpe2tree model usually has a higher distortion score during training, as we would expect after our previous findings from Figure 3.

**Tying Reordering and Syntax** The bpe2tree model generates translations with their constituency tree and their attention-derived alignments. We can use this information to extract GHKM rules (Galley et al., 2004).[6] We derive hard alignments for that purpose by treating every source/target token-pair with attention score above 0.5 as an alignment. Extracting rules from the dev-set predictions resulted in 233,657 rules, where 22,914 of them (9.8%) included reordering, i.e. contained variables ordered differently in the source and the target. We grouped the rules by their LHS (corresponding to a target syntactic structure), and sorted them by the total number of RHS (corresponding to a source sequential structure) with reordering. Table 3 shows the top 10 extracted LHS, together with the top-5 RHS, for each rule. The most common rule, VP($x_0$:TER $x_1$:NP) → $x_1$ $x_0$, found in 184 sentences in the dev set (8.4%), is indicating that the sequence $x_1$ $x_0$ in German was reordered to form a verb phrase in English, in which $x_0$ is a terminal and $x_1$ is a noun phrase. The extracted GHKM rules reveal very sensible German-English reordering patterns.

---

[5] We also note that in bins 4-6 the bpe2bpe model had slightly more translations, but this was not consistent among different runs, unlike the gaps in bins 0-3 which were consistent and contain most of the translations.

[6] github.com/joshua-decoder/galley-ghkm

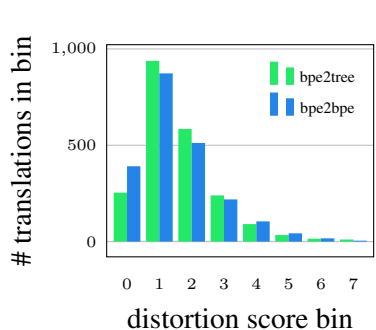
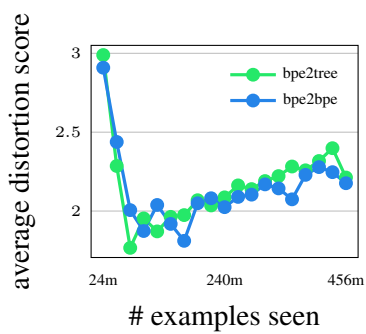
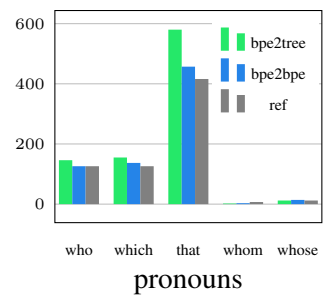

Figure 3: newstest2015 DE-EN translations binned by distortion amount

Figure 4: Average distortion score on the dev-set during different training stages

Figure 5: Amount of English relative pronouns in newstest2015 translations

| LHS | Top-5 RHS, sorted according to count. |
|---|---|
| VP($x0$:TER x1:NP) | (244) x0 x1    (157) **x1 x0**    (80) x0 x1 ",/."    (56) **x1 x0 ",/."**    (17) x0 "eine" x1 |
| VP(x0:TER PP(x1:TER x2:NP)) | (90) **x1 x2 x0**    (65) x0 x1 x2    (31) **x1 x2 x0 ",/."**    (13) x0 x1 x2 ",/."    (7) **x1 "der" x2 x0** |
| VP(x0:TER x1:PP) | (113) **x1 x0**    (82) x0 x1    (38) **x1 x0 ",/."**    (18) x0 x1 ",/."    (5) ",/." x0 x1 |
| S(x0:NP VP(x1:TER x2:NP)) | (69) x0 x1 x2    (51) **x0 x2 x1**    (35) x0 x1 x2 ",/."    (20) **x0 x2 x1 ",/."**    (6) "die" x0 x1 x2 |
| VP(x0:TER x1:NP x2:PP) | (52) x0 x1 x2    (38) **x1 x2 x0**    (20) **x1 x2 x0 ",/."**    (11) x0 x1 x2 ",/."    (9) **x2 x1 x0** |
| VP(x0:TER x1:NP PP(x2:TER x3:NP)) | (40) x0 x1 x2 x3    (32) **x1 x2 x3 x0**    (18) **x1 x2 x3 x0 ",/."**    (8) x0 x1 x2 x3 ",/."    (5) **x2 x3 x1 x0** |
| VP(x0:TER NP(x1:NP x2:PP)) | (61) x0 x1 x2    (38) **x1 x2 x0**    (19) x0 x1 x2 ",/."    (8) x0 "eine" x1 x2    (8) **x1 x2 x0 ",/."** |
| NP(x0:NP PP(x1:TER x2:NP)) | (728) x0 x1 x2    (110) "die" x0 x1 x2    (107) x0 x1 x2 ",/."    (56) x0 x1 "der" x2    (54) "der" x0 x1 x2 |
| S(VP(x0:TER x1:NP)) | (41) **x1 x0**    (26) x0 x1    (14) **x1 x0 ",/."**    (7) "die" **x1 x0**    (5) x0 x1 ",/." |
| VP(x0:TER x1:VP) | (73) x0 x1    (38) **x1 x0**    (25) x0 x1 ",/."    (15) **x1 x0 ",/."**    (9) ",/." x0 x1 |

Table 3: Top dev-set GHKM Rules with reordering. Numbers: rule counts. Bolded: reordering rules.

| | | | |
|---|---|---|---|
| src | Dutzende türkischer Polizisten wegen "Verschwörung" gegen die Regierung festgenommen | | |
| ref | Tens of Turkish Policemen Arrested over 'Plotting' against Gov't | | |
| 2tree | dozens of Turkish police arrested for "conspiracy" against the government. | | |
| 2bpe | dozens of turkish policemen on "conspiracy" against the government arrested | | |
| src | Die Menschen in London weinten, als ich unsere Geschichte erzählte. | | Er ging einen Monat nicht zu Arbeit. |
| ref | People in London were crying when I told our story. | | He ended up spending a month off work. |
| 2tree | the people of london wept as I told our story. | | he did not go to work a month. |
| 2bpe | the people of London, when I told our story. | | he went one month to work. |
| src | Achenbach habe für 121 Millionen Euro Wertgegenstände für Albrecht angekauft. | | |
| ref | Achenbach purchased valuables for Albrecht for 121 million euros. | | |
| 2tree | Achenbach has bought valuables for Albrecht for 121 million euros. | | |
| 2bpe | Achenbach have purchased value of 121 million Euros for Albrecht. | | |
| src | Apollo investierte 2008 1 Milliarde $ in Norwegian Cruise. | | Könntest du mal mit dem "ich liebe dich" aufhören? |
| ref | Apollo made a $1 billion investment in Norwegian Cruise in 2008. | | Could you stop with the "I love you"? |
| 2tree | Apollo invested EUR $1 billion in Norwegian Cruise in 2008. | | Could you stop saying "I love you"? |
| 2bpe | Apollo invested 2008 $1 billion in Norwegian Cruise. | | Can you say with the "I love you" stop? |
| src | Gerade in dieser schweren Phase hat er gezeigt, dass er für uns ein sehr wichtiger Spieler ist", konstatierte Barisic. | | |
| ref | Especially during these difficult times, he showed that he is a very important player for us", Barisic stated. | | |
| 2tree | Especially at this difficult time he has shown that he is a very important player for us," said Barisic. | | |
| 2bpe | It is precisely during this difficult period that he has shown us to be a very important player, "Barisic said. | | |
| src | Hopfen und Malz - auch in China eine beliebte Kombination. | | "Ich weiß jetzt, dass ich das kann - prima!" |
| ref | Hops and malt - a popular combination even in China. | | "I now know that I can do it - brilliant!" |
| 2tree | Hops and malt - a popular combination in China. | | "I now know that I can do that! |
| 2bpe | Hops and malt - even in China, a popular combination. | | I know now that I can that - prima!" |
| src | Die Ukraine hatte gewarnt, Russland könnte auch die Gasversorgung für Europa unterbrechen. | | |
| ref | Ukraine warned that Russia could also suspend the gas supply to Europe. | | |
| 2tree | Ukraine had warned that Russia could also stop the supply of gas to Europe. | | |
| 2bpe | Ukraine had been warned, and Russia could also cut gas supplies to Europe. | | |
| src | Bis dahin gab es in Kollbach im Schulverband Petershausen-Kollbach drei Klassen und in Petershausen fünf. | | |
| ref | Until then, the school district association of Petershausen-Kollbach had three classes in Kollbach and five in Petershausen. | | |
| 2tree | until then, in Kollbach there were three classes and five classes in Petershausen. | | |
| 2bpe | until then there were three classes and in Petershausen five at the school board in Petershausen-Kollbach. | | |

Table 4: Translation examples from newstest2015. The underlines correspond to the source word attended by the first opening bracket (these are consistently the main verbs or structural markers) and the target words this source word was most strongly aligned to. See the supplementary material for an attention weight matrix example when predicting a tree (Figure 6) and additional output examples.

**Relative Constructions** Browsing the produced trees hints at a tendency of the syntax-aware model to favor using relative-clause structures and subordination over other syntactic constructions (i.e., "several cameras *that* are all priced..." vs. "several cameras, all priced..."). To quantify this, we count the English relative pronouns (who, which, that[7], whom, whose) found in the newstest2015 translations of each model and in the reference translations, as shown in Figure 5. The bpe2tree model produces more relative constructions compared to the bpe2bpe model, and both models produce more such constructions than found in the reference.

**Main Verbs** While not discussed until this point, the generated opening and closing brackets also have attention weights, providing another opportunity to to peak into the model's behavior. Figure 6 in the supplementary material presents an example of a complete attention matrix, including the syntactic brackets. While making full sense of the attention patterns of the syntactic elements remains a challenge, one clear trend is that opening the very first bracket of the sentence *consistently attends to the main verb or to structural markers* (i.e. question marks, hyphens) in the source sentence, suggesting a planning-ahead behavior of the decoder. The underlines in Table 4 correspond to the source word attended by the first opening bracket, and the target word this source word was most strongly aligned to. In general, we find the alignments from the syntax-based system more sensible (i.e. in Figure 1 – the bpe2bpe alignments are off-by-1).

---

[7]"that" also functions as a determiner. We do not distinguish the two cases.

**Qualitative Analysis and Human Evaluations**
The bpe2tree translations read better than their bpe2bpe counterparts, both syntactically and semantically, and we highlight some examples which demonstrate this. Table 4 lists some representative examples, highlighting improvements that correspond to syntactic phenomena involving reordering or global structure. We also performed a small-scale human-evaluation using mechanical turk on the first 500 sentences in the dev-set. Further details are available in the supplementary material. The results are summarized in the following table:

| | |
|---|---|
| 2bpe weakly better | 100 |
| 2bpe strongly better | 54 |
| 2tree weakly better | 122 |
| 2tree strongly better | 64 |
| both good | 26 |
| both bad | 3 |
| disagree | 131 |

As can be seen, in 186 cases (37.2%) the human evaluators preferred the bpe2tree translations, vs. 154 cases (30.8%) for bpe2bpe, with the rest of the cases (30%) being neutral.

## 4 Conclusions and Future Work

We present a simple string-to-tree neural translation model, and show it produces results which are better than those of a neural string-to-string model. While this work shows syntactic information about the target side can be beneficial for NMT, this paper only scratches the surface with what can be done on the subject. First, better models can be proposed to alleviate the long sequence problem in the linearized approach or allow a more natural tree decoding scheme (Alvarez-Melis and Jaakkola, 2017). Comparing our approach to other syntax aware NMT models like Eriguchi et al. (2017) and Nadejde et al. (2017) may also be of interest. A Contrastive evaluation (Sennrich, 2016) of a syntax-aware system vs. a syntax-agnostic system may also shed light on the benefits of incorporating syntax into NMT.

## Acknowledgments

This work was supported by the Intel Collaborative Research Institute for Computational Intelligence (ICRI-CI), and The Israeli Science Foundation (grant number 1555/15).

# A   Supplementary Material

**Data**   The English side of the corpus was tokenized (into Penn treebank format) and truecased using the scripts provided in Moses (Koehn et al., 2007). We ran the BPE process on a concatenation of the source and target corpus, with 89500 BPE operations in the WMT experiment and with 45k operations in the other experiments. This resulted in an input vocabulary of 84924 tokens and an output vocabulary of 78499 tokens in the WMT16 experiment. The linearized constituency trees are obtained by simply replacing the POS tags in the parse trees with the corresponding word or subwords. The output vocabulary in the bpe2tree models includes the target subwords and the tree symbols which correspond to an opening or closing of a specific phrase type.

**Hyperparameters**   The word embedding size was set to 500/256 and the encoder and decoder sizes were set to 1024/256 (WMT16/other experiments). For optimization we used Adadelta (Zeiler, 2012) with minibatch size of 40. For decoding we used beam search with a beam size of 12. We trained the bpe2tree WMT16 model on sequences with a maximum length of 150 tokens (the average length for a linearized tree in the training set was about 50 tokens). It was trained for two weeks on a single Nvidia TitanX GPU. The bpe2bpe WMT16 model was trained on sequences with a maximum length of 50 tokens, and with minibatch size of 80. It was trained for one week on a single Nvidia TitanX GPU. Only in the low-resource experiments we applied dropout as described in Sennrich et al. (2016a) for Romanian-English.

**Human Evaluation**   We performed human-evaluation on the Mechnical Turk platform. Each sentence was evaluated using two annotators. For each sentence, we presented the annotators with the English reference sentence, followed by the outputs of the two systems. The German source was not shown, and the two system's outputs were shown in random order. The annotators were instructed to answer "Which of the two sentences, in your view, is a better portrayal of the the reference sentence." They were then given 6 options: "sent 1 is better", "sent 2 is better", "sent 1 is a little better", "sent 2 is a little better", "both sentences are equally good", "both sentences are equally bad". We then ignore differences between "better" and "a little better". We count as "strongly better" the cases where both annotators indicated the same sentence as better, as "weakly better" the cases were one annotator chose a sentence and the other indicated they are both good/bad. Other cases are treated as either "both good" / "both bad" or as disagreements.

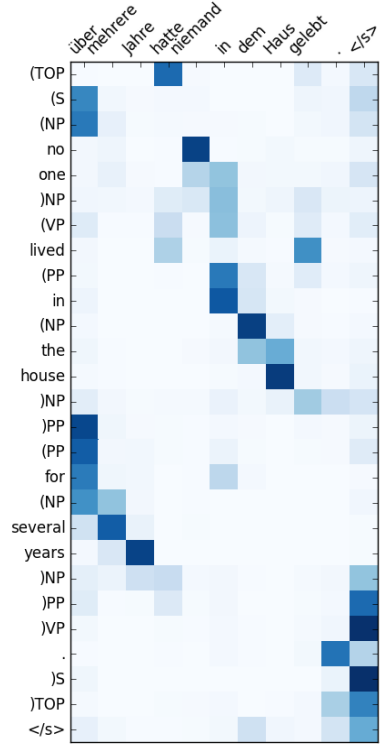

Figure 6: The attention weights for the string-to-tree translation in Figure 1

**Additional Output Examples**   from both models, in the format of Figure 1. Notice the improved translation and alignment quality in the tree-based translations, as well as the overall high structural quality of the resulting trees. The few syntactic mistakes in these examples are attachment errors of SBAR and PP phrases, which will also challenge dedicated parsers.

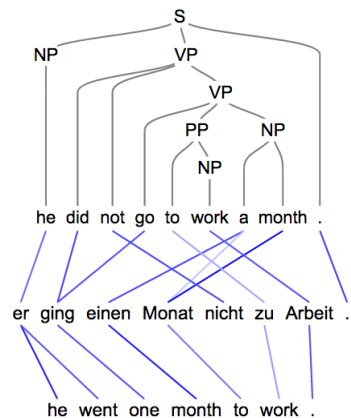

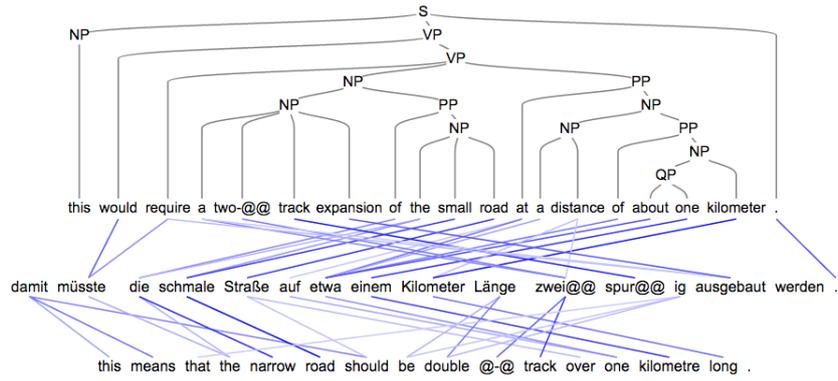

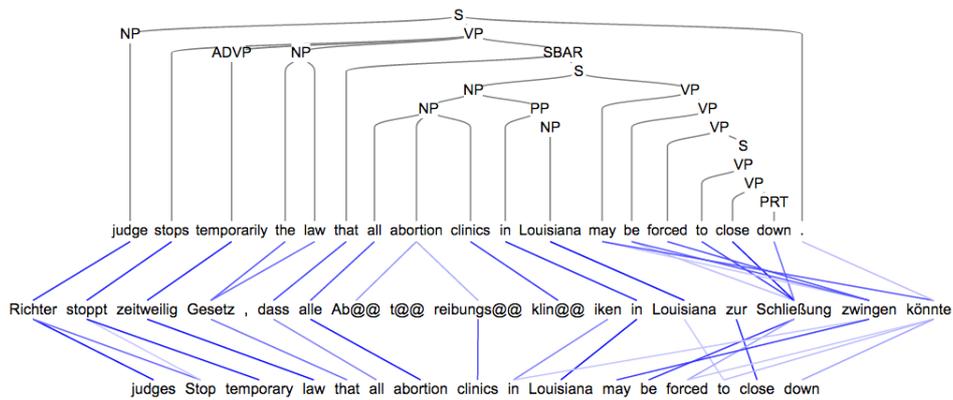

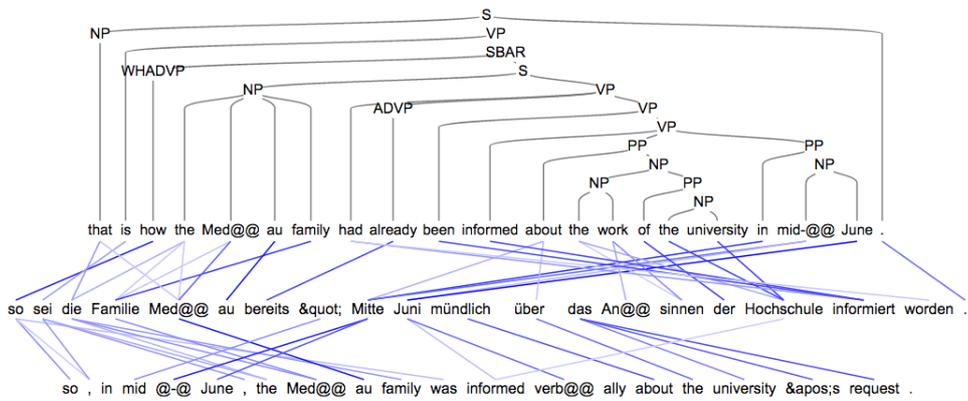

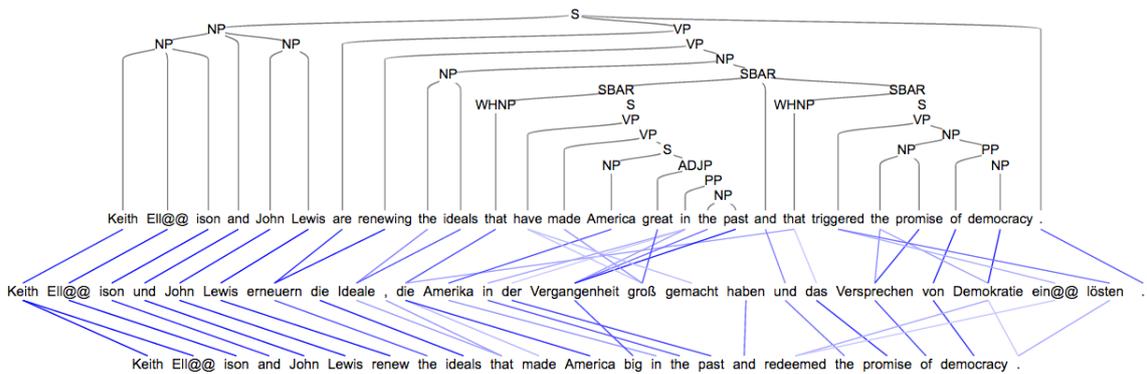